\documentclass[manuscript]{acmart}
\AtBeginDocument{%
  \providecommand\BibTeX{{%
    \normalfont B\kern-0.5em{\scshape i\kern-0.25em b}\kern-0.8em\TeX}}}

\copyrightyear{2024}
\acmYear{2024}
\setcopyright{acmlicensed}\acmConference[GUIDE-AI '24]{Governance, Understanding and Integration of Data for Effective and Responsible AI}{June 14, 2024}{Santiago, AA, Chile}
\acmBooktitle{Governance, Understanding and Integration of Data for Effective and Responsible AI (GUIDE-AI '24), June 14, 2024, Santiago, AA, Chile}
\acmDOI{10.1145/3665601.3669841}
\acmISBN{979-8-4007-0694-3/24/06}

\usepackage{subcaption}
\begin{document}

\title{TbExplain: A Text-Based Explanation Method for Scene Classification Models With the Statistical Prediction Correction}

\author{Amirhossein Aminimehr}
\affiliation{%
  \institution{Iran University of Science and Technology}
  \city{Tehran}
  \country{Iran}}
\email{Amir_aminimehr@comp.iust.ac.ir}

\author{Pouya Khani}
\authornote{Both authors contributed equally to this research.}
\affiliation{%
  \institution{Aarhus University}
  \city{Aarhus}
  \country{Denmark}}
\email{pouya.khani@cs.au.dk}

\author{Amirali Molaei}
\authornotemark[1]
\affiliation{%
  \institution{Iran University of Science and Technology}
  \city{Tehran}
  \country{Iran}}
\email{amirali_molaei@comp.iust.ac.ir}

\author{Amirmohammad Kazemeini}
\affiliation{%
  \institution{Nanyang Technological University}
  \country{Singapore}}
\email{akazemei@uwo.ca}

\author{Eric Cambria}
\affiliation{%
  \institution{Nanyang Technological University}
  \country{Singapore}}
\email{cambria@ntu.edu.sg}
\renewcommand{\shortauthors}{Aminimehr, et al.}

\begin{abstract}
Heatmaps are common tools in Explainable Artificial Intelligence (XAI) field, but they are not without imperfections; E.g., non-expert users may not grasp the underlying rationale of heatmaps, wherein pixels relevant to the model's prediction are highlighted through distinct intensities or colors. Moreover, objects and regions of the input image that are relevant to the model prediction are frequently not entirely differentiated by heatmaps. In this paper, we propose a framework called TbExplain that employs XAI techniques and a pre-trained object detector to present text-based explanations of scene classification models. Moreover, TbExplain incorporates a novel method to correct predictions and textually explain them based on the statistics of objects in the input image when the initial prediction is unreliable. To assess the trustworthiness and validity of the text-based explanations, we conducted a qualitative experiment, and the findings indicated that these explanations are sufficiently reliable.
Furthermore, our quantitative and qualitative experiments on TbExplain with scene classification datasets reveal an improvement in classification accuracy over ResNet variants.
\end{abstract}





\maketitle

\section{Introduction}
Several state-of-the-art architectures, such as ResNet~\cite{he2016deep}, have progressed to the point where they even surpass human-level performance in image classification task~\cite{he2015delving}. Deep Neural Networks (DNNs) are over-parameterized and black-box; Thus, it is often difficult to decipher how predictions are generated by them~\cite{doshi2017towards}. This challenge is associated with "lack of interpretability/explainability" issue that prevents the adoption of DNNs in sensitive applications due to lack of trust in their output generation process~\cite{carvalho2019machine}. The lack of interpretability of DNNs, restricts their users from comprehending the reasoning behind their predictions, resulting in a decline in their usage~\cite{meng2020interpreting,figueroa2021towards}. 

The generation of a saliency map~\cite{simonyan2013deep} (also known as \textit{heatmap}) on the model's input features is one of the most often employed Explainable Artificial Intelligence (XAI) approaches for deep models. Each pixel in a heatmap represents the contribution of a specific feature to the model's prediction, with color intensity indicating the magnitude: green for positive contributions (pushing towards a positive class or higher probability) and red for negative contributions (pushing towards a negative class or lower probability). A saliency map facilitates the evaluation of input feature importance in a model's prediction by highlighting relevant regions. However, its interpretability may pose challenges for non-expert users (users unfamiliar with the heatmap's rationale), as they may struggle to discern the correlation between input features and the model's output. Moreover, heatmaps often fail to distinctly delineate objects and regions crucial to the prediction, resulting in fragmented and dispersed representations that hinder accurate user recognition within the associated input. This challenge is particularly pronounced in human-computer interaction applications, where comprehending visual information is already difficult for individuals with visual impairments, thus reducing accessibility for a wider audience ~\cite{cho2021multi}.

In this paper, we introduce a novel method to explain scene classification models through textual explanations. In contrast to heatmaps, this approach offers a clear textual representation, addressing issues of user interpretability and avoiding dispersion of relevant regions. Additionally, the method improves classification performance by implementing a confidence measure and a logic-based decision maker. The contribution and novelty of this work are summarized as follows:

\begin{itemize}
	\item
	We propose a novel framework designed to offer context-based textual explanations with the goal of improving the interpretability of the underlying mechanisms of scene classification models for both expert and non-expert users.
	
	\item We present a heuristic-based method called statistical prediction correction (SPC) to obtain a textually explainable revised prediction based on the statistics of objects in the input image when the original prediction is unreliable.
 
	\item Through qualitative analysis, we show that text-based explanations generated by TbExplain are reliable and trustworthy. Also, through quantitative experiments, we show that TbExplain can increase the accuracy of scene classification models.
\end{itemize}

\section{Related works}
\label{related}
As our research focuses on generating a text-based explanation for scene classification models based on saliency-based XAI methods, in some way, our work is related to articles that use saliency-based XAI methods to visually explain an image classification model (or other tasks in which their inputs are images) \cite{sundararajan2017axiomatic,lundberg2017unified}. LIME~\cite{ribeiro2016should} and GradCAM~\cite{selvaraju2017grad}, that we used them in our proposed method, are two examples of using saliency-based XAI techniques for image classification task too.

Our research is also similar to the image captioning task, which takes an image as input and generates textual narratives that describe the image at hand in an encoder-decoder fashion~\cite{stefanini2022show}. Many research efforts have been put into developing an image captioning system capable of explaining visual data descriptively in a text-based format~\cite{vinyals2015show,rennie2017self,huang2019attention}.



TbExplain can be considered to be more related to the following studies in terms of the use of environmental objects as important components to improve scene classification. Chen et al.~\cite{chen2019scene} considered objects as the context in the scene and proposed to derive object embeddings by implementing an object segmentation module, and used these vectors to refine the top five predictions of ResNet~\cite{he2016deep} to improve scene recognition. On the other hand, Heikel et al.~\cite{heikel2022indoor} used YOLO as an object detection architecture to extract objects from the scene.  They further mapped these objects to TF-IDF feature vectors and used them to train a scene classifier to predict scene categories. Regarding the explainability aspect of scene classification, the work of Anjomshoae et al.~\cite{anjomshoae2021context} is most related to our approach of providing text-based explanations using local information (i.e. objects). It generated textual explanations by calculating the contextual importance of each semantic category in the scene by masking that particular segment, and determining its effect on the prediction. In contrast to these methods, our framework does not solely rely on the scene objects to classify images; instead, it uses their information as a modifier to correct the output of the scene classification module when its softmax score does not exceed a certain threshold.

\section{Proposed method}
\label{proposed}
\begin{figure*}
	\centering
	\includegraphics[width=1\textwidth, height=0.25\textheight]{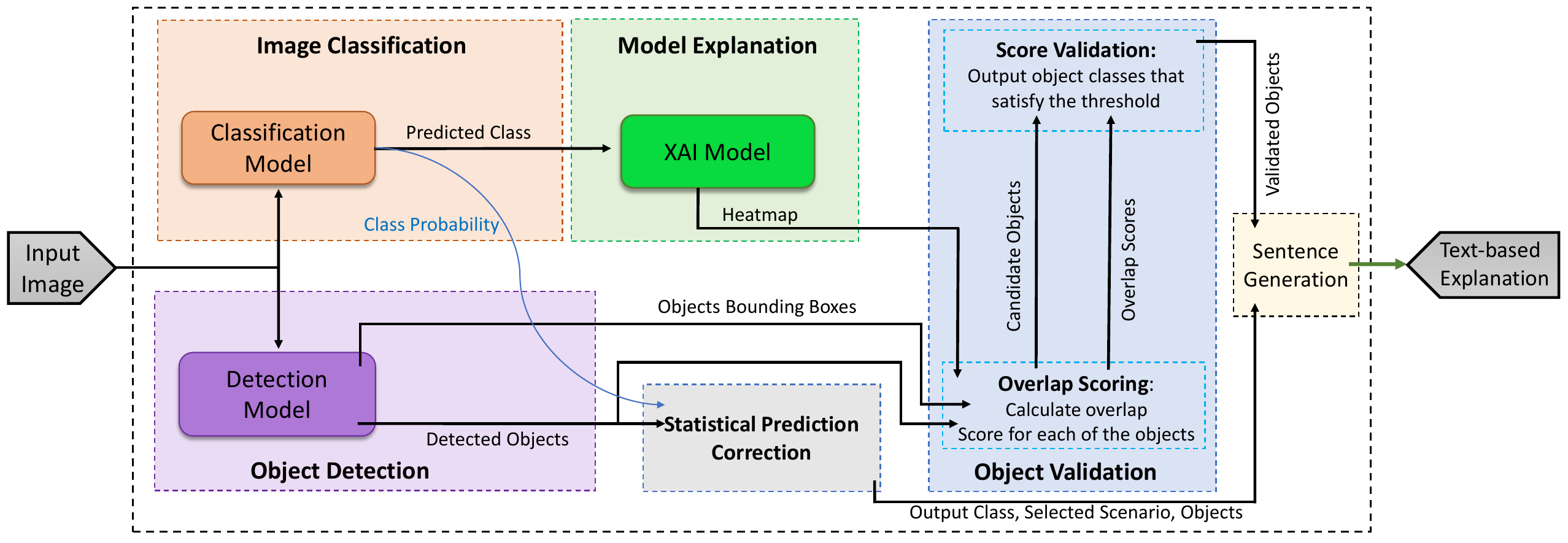}
	\caption{An overview of the proposed framework.}
	\label{Frame}
\end{figure*}

Figure \ref{Frame} shows our proposed method, called TbExplain. TbExplain includes six major modules: (1) Image Classification, (2) Object Detection, (3) Model Explanation, (4) Object Validation, (5) Statistical Prediction Correction, and (6) Sentence Generation. In the image classification and object detection modules, we used pre-trained classifier and detector, respectively. In model explanation module, we used several saliency-based XAI methods to generate explanation heatmaps. Following is a concise description of other modules that make up the framework.

\subsection{Object validation}
To determine which objects contribute to the generation of the output of the image classification model, we propose a novel technique to do so by assessing a fraction called \textit{Overlapping Score} ($OS$). The overlapping score for each object reflects the percentage of the bounding box covered by the overall heatmap. For example, a score of $OS = 100$ indicates that the heatmap extends throughout the boundary box, and a score of 0 means that the boundary box is located in a place where it does not have an intersection with the heatmap.
Given the generated heatmap (\emph{X}), as well as the set of detected bounding boxes (\emph{B}), the process of determining the classes $\{c_1,c_2,…,c_i\}$ that exceed a certain threshold can be formulated as:

\begin{equation}
	\{c_1,c_2,…,c_i\} = OV(X,B)
\end{equation}

\begin{figure*}
	\centering
	\includegraphics[width=0.8\textwidth,height=0.15\textheight]{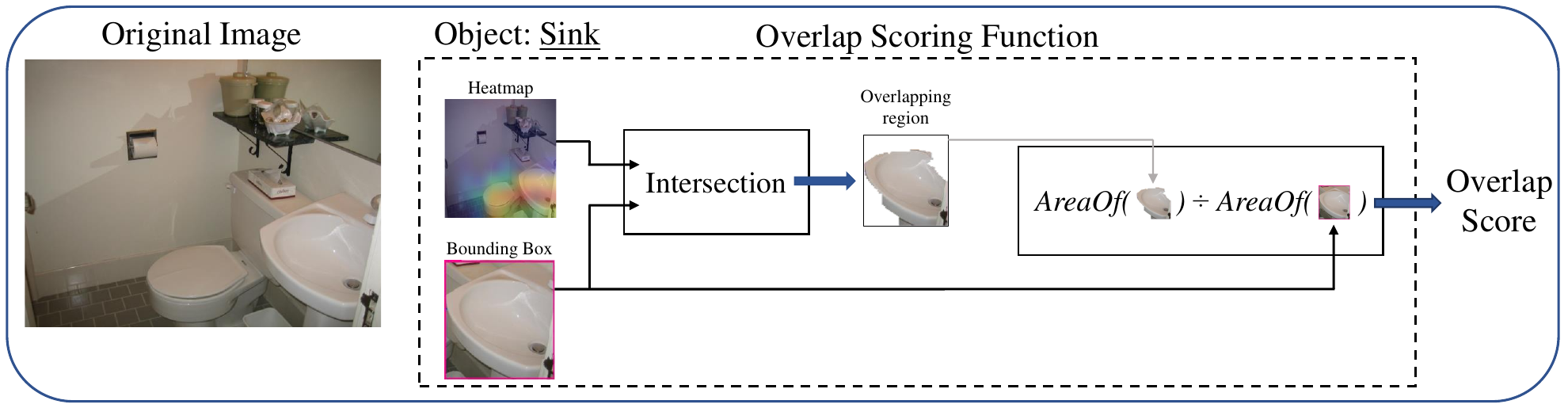}
	\caption{The structure of the overlap scoring function. $b$ is the object bounding box and $OR$ is the overlapping region.}
	\label{OSF}
\end{figure*}

where $OV(.)$ represents the Object Validation module. As depicted in figure \ref{OSF}, the overlapping region between the heatmap and the bounding box of the object \emph{i} is obtained through an intersection operation. The overlap score is then calculated by dividing the area of the overlapped region by the area of the corresponding object bounding box, which is a number between 0 and 1. The whole operation can be defined as follows:

\begin{equation}
	OS_i = \frac{Intersection(X,b_i)}{b_i}
\end{equation}

where $OS_i$ is the overlap score of the object \emph{i}, and $b_i$ is the boundary box of the object \emph{i}.
The object classes $\{c_1,c_2,…,c_i\}$ that make significant contributions to the scene classification process are obtained through the scoring validation function. Additionally, the detection confidence of the objects $DC_i$ will be multiplied by $OS_i$ to obtain a relevance score $RS_i$. Finally, if $RS_i$ exceeds a relevance threshold $T_R$ for the corresponding object $i$, the module will output the class of object $c_i$ as the validated object. A high relevence score indicates that not only is the object detected correctly, but it also contributes significantly to the classification.

\subsection{Statistical prediction correction}
Statistical Prediction Correction (SPC) is a logic-based decision maker conditioned on a confidence threshold, the softmax probability of the output class, to determine if the revision of the scene classification prediction is required and its procedure should be executed. SPC statistically learns $N_c$ (number of classes) weights for each object in the training phase by incrementally calculating the probability of the object occurring in each class:

\begin{equation}
	W_{ic} = \frac{d_{ic}}{D_i}
\end{equation}

Where $W_{ic}$ is denoted as the corresponding weight of an object \emph{i} for class \emph{c}, $d_{ic}$ as the number of times the object \emph{i} has occurred in scenes belonging to class \emph{c}, and $D_i$ as the quantification of the object \emph{i} that occurs in all scenes of the data set. It is important to note that these weight assessments are intended for objects validated by the Object Validation module. In the test phase, if the scene classification model's prediction can't be regarded as reliable, the prediction is entrusted to the SPC, which leverages the estimated weights to decide the category of the input via calculating a weighted score for each of the classes in the following manner:
\begin{equation}
    s_c = n_{1}W_{1c} + n_{2}W_{2c} + \cdots + n_{i}W_{ic}
\end{equation}
where $n_{i}$ indicates the quantity of object \emph{i} existing in the scene. In contrast to the training phase, every object's weight is incorporated to obtain the weighted score since the heatmap produced by the XAI technique is not accurate. Next, the class with the highest score \emph{s} wins as the label of the input. The textual explanation is provided in a constant format, with only the predicted class and the intended objects changing with respect to the input.

\subsubsection{The first scenario}

The first scenario occurs when the probability value obtained for the predicted class is higher than a specific threshold, indicating that the model has made its prediction with enough confidence. When this happens, the input image and the classification model are fed into the proposed method to generate a text-based explanation, as demonstrated in figure \ref{fig}. In this scenario, the textual explanation is provided by representing the objects that contributed most to the model's classification, and the output phrase confirms the class of the input by referring to these validated objects as the reason for the process.

\subsubsection{The second scenario}
The second scenario occurs when the probability value is below a predetermined threshold, indicating an unreliable prediction by the scene classification model. In this case, the SPC corrects the output class by generating a revised label based on the objects present in the input image. As a result, this approach outputs items that are relevant to the new predicted label. Figure \ref{fig} visually illustrates the output generated by the SPC approach in the second scenario.

\subsubsection{The third scenario}
\begin{figure*}
	\centering
	\includegraphics[width=0.65\textwidth, height=0.3\textheight]{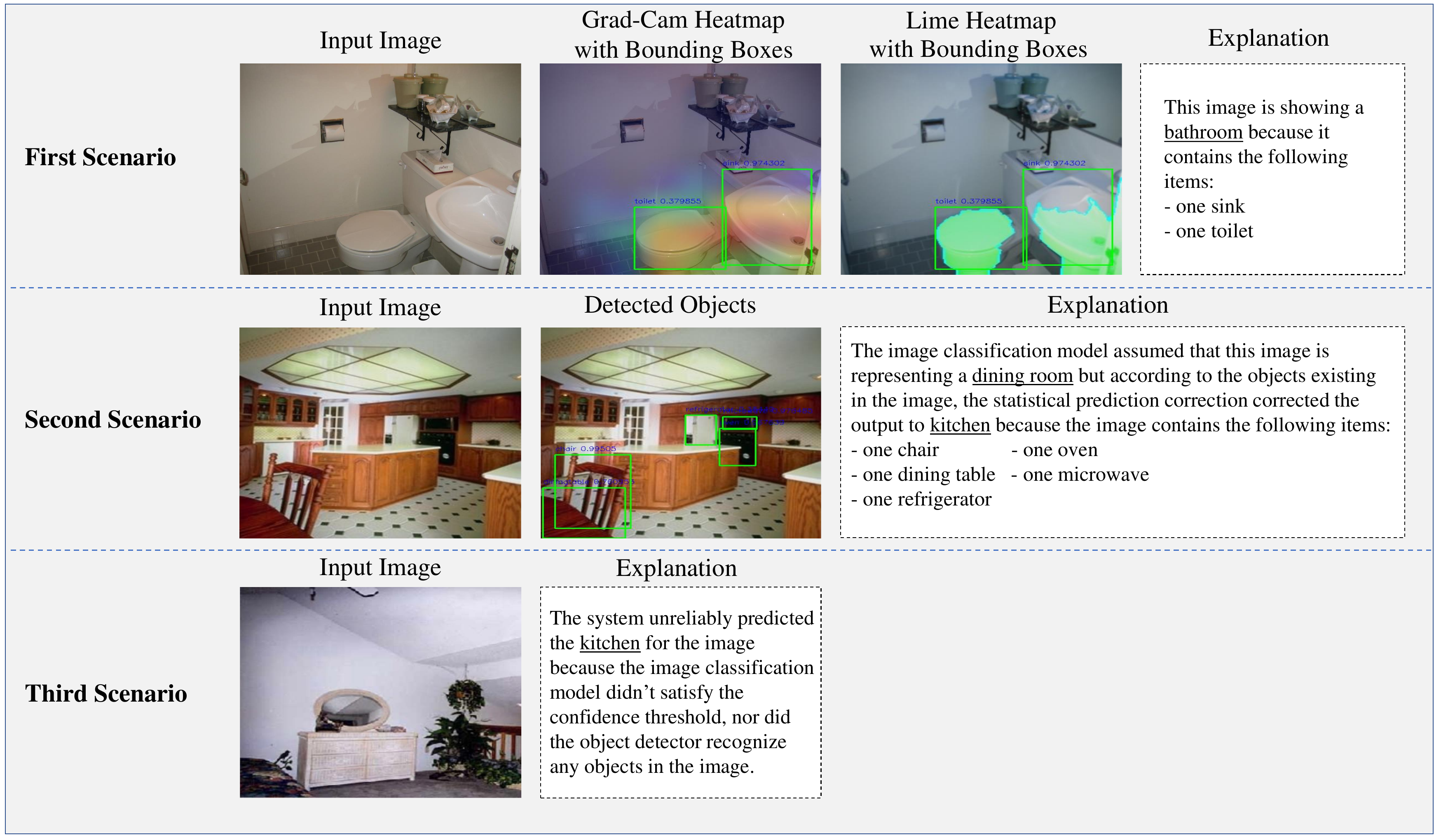}
	\caption{Outputs of TbExplain in three scenarios.}
	\label{fig}
\end{figure*}

The third scenario arises when the image classification model's probability value for the top class falls below a predetermined threshold and the object detection module fails to detect any objects in the input image. In such cases, the interpretation highlights the unreliability of the system in making accurate predictions and displays the output of the image classification model. Moreover, the SPC module is unable to correct the unreliable predictions generated by the scene classification model. The output of TbExplain in this scenario is depicted in figure \ref{fig} and suggests that the model may require further refinement and diagnosis.

\subsection{Sentence generation}

\begin{figure*}
	\centering	\includegraphics[width=0.65\textwidth, height= 0.28\textheight]{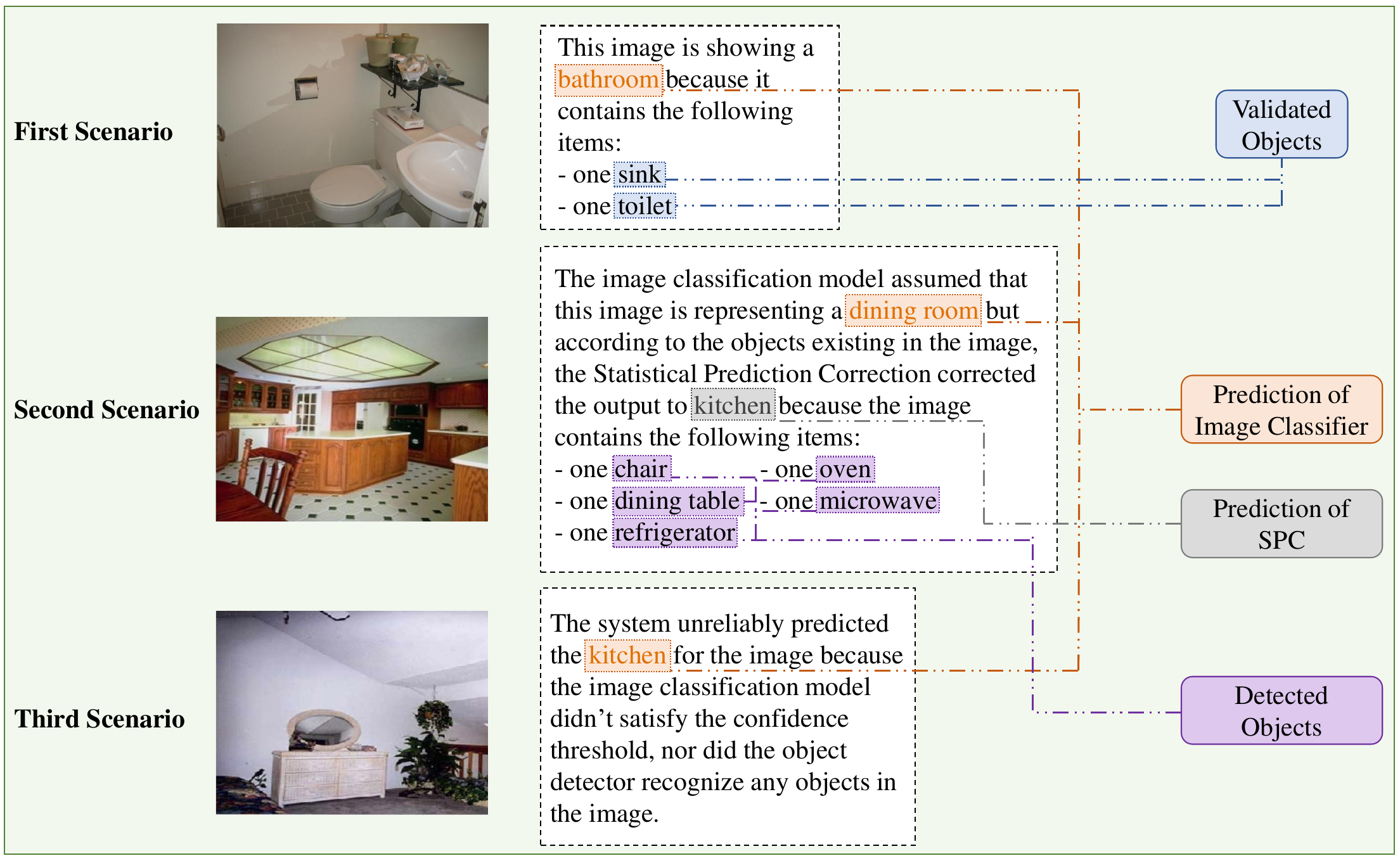}
	\caption{Examples of the sentence generation's outputs (i.e., the text-based explanation).}
	\label{SG}
\end{figure*}

To output the desired textual explanation, we implemented the Sentence Generation module to leverage the information provided by both the SPC and the object validation modules and integrate them to set up the explanatory sentence. The sentence generation procedure can be interpreted as a fill-in-the-blank task. In each scenario, a fixed sentence with several missing words that are either the prediction or the pertaining objects is the general structure of the explanation. As illustrated in Figure \ref{SG}, the objective is to provide this missing information by utilizing the scene classification's prediction in all scenarios, the validated objects in the first scenario, the revised prediction of the SPC module, and all detected objects in the second scenario.

\section{Experiments}
\label{experiment}
 
We selected three datasets to evaluate the performance of TbExplain: MIT67 \cite{quattoni2009recognizing}, Places365 \cite{zhou2017places}, and SUN397 \cite{xiao2010sun}. Due to the vast size of datasets, the hardware limitations for processing these data, and the considerable similarity of a number of classes to one another, nine unique categories out of all categories were selected for each dataset, followed by the selection of all images labeled with one of these nine classes.
We randomly selected 78 images per chosen category from the MIT67 dataset and 75 images for the other two datasets as training data. These new datasets, derived from the original datasets and each with nine categories, were selected as our experimental data. To generate test data for each dataset, we randomly selected 25 images for each of the nine classes. We also randomly selected 10\% of the training data as validation data. Finally, we convert all image sizes to $256\times256$ pixel and normalize all pixel values from $[0,255]\in \mathbb{N}$ to $[0,1]\in \mathbb{R}$. We utilized various versions of pre-trained ResNet on the ImageNet dataset~\cite{russakovsky2015imagenet} and a pre-trained YOLOv5 on the COCO dataset~\cite{lin2014microsoft} as the scene classification and object detection models, respectively. YOLOv5 is used beacause it incorporates a range of techniques and architectural innovations aimed at enhancing its robustness to noise. These include a resilient network architecture, effective post-processing, and robust training techniques, all contributing to its ability to maintain high detection accuracy in noisy environments. We also employed GradCAM and LIME as internal XAI techniques in the Model Explanation module. The computational resources for this study are a 12GB RAM and a Nvidia Tesla K80 GPU.

\subsection{Qualitative analysis}
In this section, we evaluate the reliability of TbExplain's textual explanations by comparing its results with the manually annotated data. In detail, 108 scene images from nine different classes (twelve images from each class) were randomly divided among the three authors of this paper to manually annotate them. In fact, each author should look at each scene and identify and record three of the most relevant and important objects in the scene that are representative of that class (e.g., the "bed" object is a representative of the "bedroom" class). If less than three representative objects were recognizable in the image, the author should fill the empty slots with "empty".

Finally, we compare the identified objects with the objects presented in the TbExplain textual explanation for each scene. We obtained 65.94\% accuracy in this experiment. It means that in 65.94\% of cases, our method reliably and correctly interprets the reasoning involved in scene classification by presenting the most relevant objects within the scene (as representative objects of that class) in its textual explanation.

\subsection{Quantitative analysis}

\begin{figure}[t!]
	\centering	\includegraphics[width=0.95\textwidth,height=0.28\textheight]{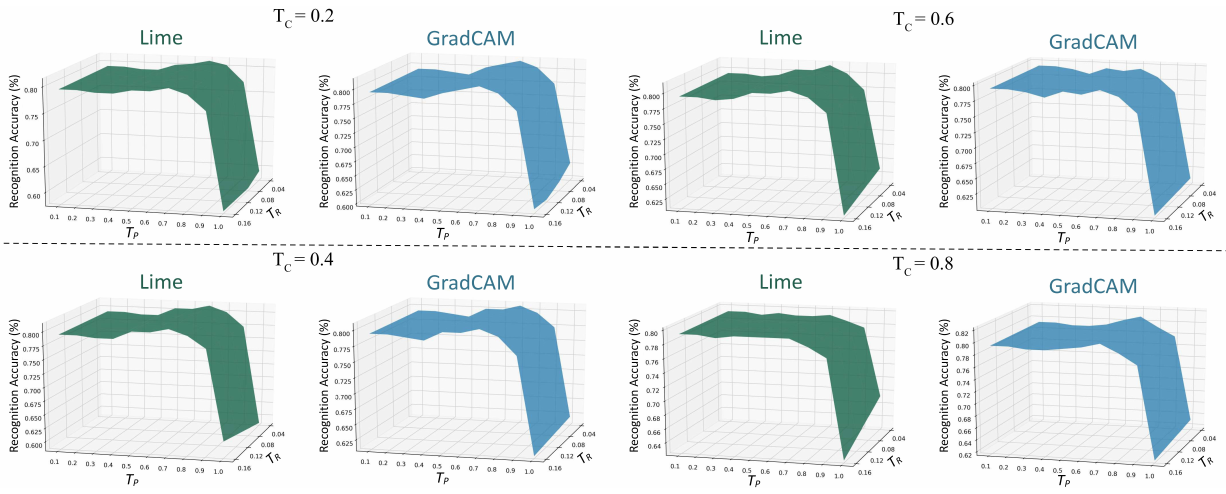}
	\caption{Recognition accuracy of TbExplain on the validation set based on the specified thresholds $T_C$, $T_R$, and $T_P$.}
	\label{thresh}
\end{figure}

In this section, we quantitatively evaluate the effectiveness of TbExplain (in the second scenario using the SPC method) by comparing its performance with the original image classification model's accuracy. For tuning the proposed method's performance, we should adjust three hyperparameters:
\begin{enumerate}
	\item \textit{Confidence threshold} ($T_C$), which specifies in the object detection module whether a detected object is a valid object or not.
	
	\item \textit{Relevance threshold} ($T_R$), which specifies whether a valid object is a relevant object according to the predicted class or not.
	
	\item \textit{Class Probability threshold} ($T_P$) which is a threshold on the probability value of the top class in the last layer of the scene classification model that specifies the scenarios mentioned in the previous section. In fact, $T_P$ is a threshold on the model prediction confidence.
\end{enumerate}

We employed grid-search with 4-fold cross-validation on the MIT67 dataset and validated TbExplain to assess the best values of the thresholds based on the validation data. The grid search procedure is performed using the following parameters: values of $\{0.2, 0.4, 0.6, 0.8\}$ for $T_C$, values of $\{0.2T_C, 0.4T_C, 0.6T_C, 0.8T_C\}$ for $T_R$, and 10 distinct values for $T_P$, from $0.1$ to $1$ with a step size of $0.1$. 

Figure \ref{thresh} demonstrates grid-search results on the validation data with $T_C$, $T_R$, and $T_P$ as the parameters. The best performance is achieved with $T_C=0.2, T_R=0.08$, and $T_P=0.7$. The comparison of TbExplain on the test set with various variations of ResNet represents an improvement in terms of prediction accuracy, with Lime-integrated TbExplain achieving 82.47\% accuracy (4.13\% higher than the best variation of ResNet). In other words, the SPC method corrects 4.13\% of the false predictions. Table \ref{tab:tab1} summarizes the performance achieved by four versions of ResNet and TbExplain (in the second scenario using the SPC method) in the MIT67 dataset.

It should be noted that since the XAI methods are applied to the original scene classification model and cannot independently report the classification accuracy, TbExplain is actually the output of applying the proposed second scenario to the pre-trained ResNet101V2 to improve its performance. Therefore, there is no need for training and validation again, and for this reason, the accuracy value for training and validation of TbExplain is not reported in the table mentioned above.

\begin{figure*}
	\centering	\includegraphics[width=0.45\textwidth,height=0.2\textheight]{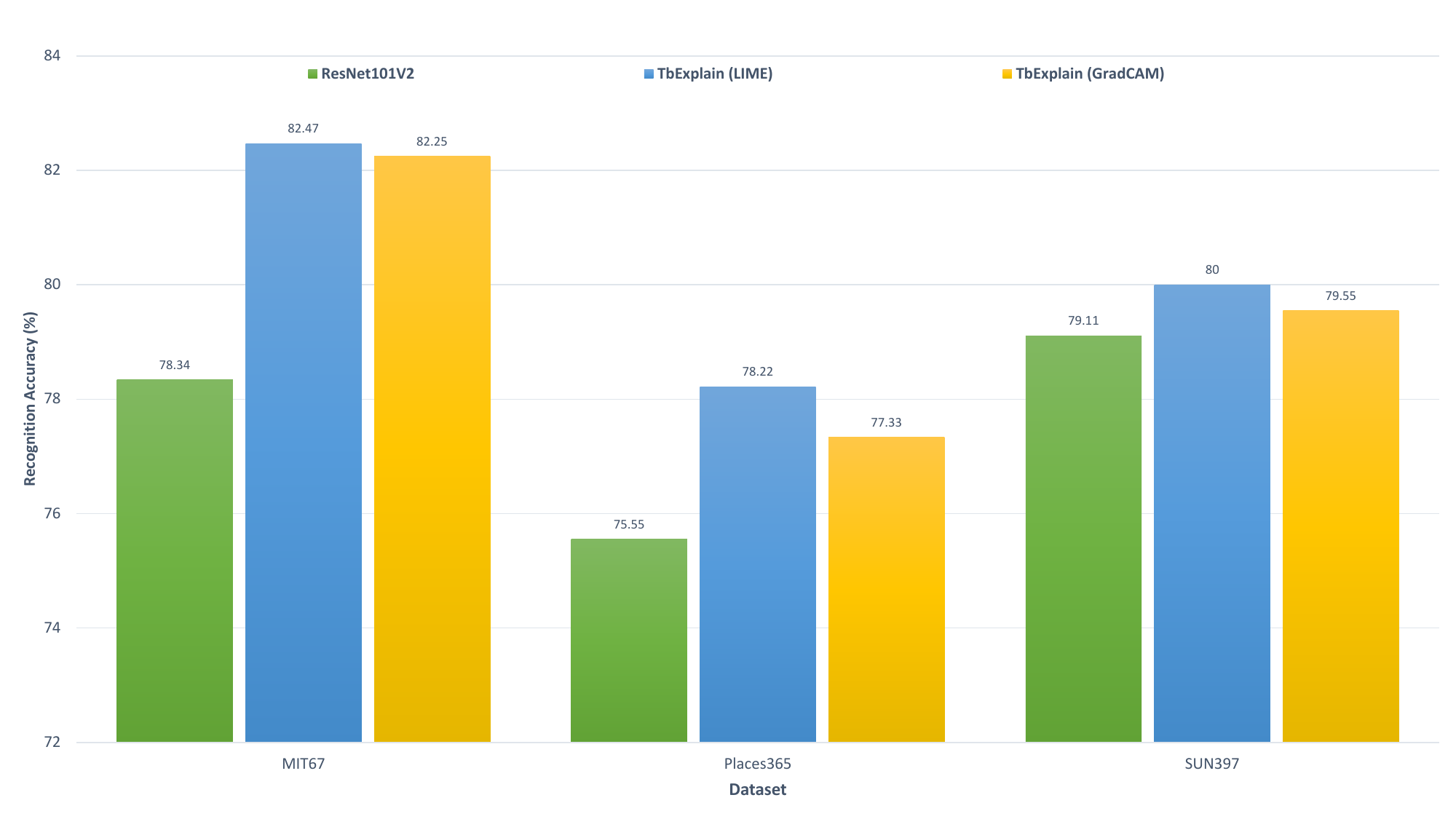}
	\caption{A summary of the performance of TbExplain and ResNet101V2 on all three datasets.}
	\label{plot}
\end{figure*}

As mentioned earlier, we also retrained all versions of ResNet on two other datasets (Places365 and SUN397) and tested TbExplain on them, utilizing the learned threshold values ($T_C=0.2, T_R=0.08$, and $T_P=0.7$) to ensure TbExplain robustness and generalizability in different scenes and environments. Since generalizing to scenes and images with different contextual information, such as different objects and backgrounds, is of high significance, we chose environments that were dissimilar to MIT67 from SUN397 and Places365. Tables \ref{tab:tab2} and \ref{tab:tab3} summarize the performance achieved by all the above-mentioned versions of ResNet and TbExplain on the Places365 and SUN397 datasets, respectively. According to these two tables, the effectiveness of our model in increasing the classification accuracy can also be seen in other datasets. Figure \ref{plot} provides a summary of the performance of TbExplain and the best performing ResNet model on all three datasets.


\begin{table}[t!]
\centering
\caption{Accuracy of TbExplain on different datasets.}
\label{tab:combined}

\begin{subtable}{.48\textwidth}
\centering
\resizebox{\linewidth}{!}{%
\begin{tabular}{c|ccc}
\hline
\textbf{Method} & \textbf{Training} & \textbf{Validation} & \textbf{Testing} \\ \hline
ResNet50 & 99.96\% & 71.07\% & 70.31\% \\
ResNet50V2 & 99.92\% & 75.35\% & 75.67\% \\
ResNet101 & 99.96\% & 71.78\% & 68.3\% \\
ResNet101V2 & 99.96\% & 79.28\% & 78.34\% \\
\textbf{TbExplain (LIME)} & \textbf{-} & \textbf{-} & \textbf{82.47\%} \\
TbExplain (GradCAM) & - & - & 82.25\% \\ \hline
\end{tabular}%
}
\caption{MIT67 dataset.}
\label{tab:tab1}
\end{subtable}
\hfill
\begin{subtable}{.48\textwidth}
\centering
\resizebox{\linewidth}{!}{%
\begin{tabular}{c|ccc}
\hline
\textbf{Method} & \textbf{Training} & \textbf{Validation} & \textbf{Testing} \\ \hline
ResNet50 & 100\% & 71.64\% & 68.44\% \\
ResNet50V2 & 100\% & 77.61\% & 74.22\% \\
ResNet101 & 100\% & 70.14\% & 68.44\% \\
ResNet101V2 & 100\% & 79.10\% & 75.55\% \\
\textbf{TbExplain (LIME)} & \textbf{-} & \textbf{-} & \textbf{78.22\%} \\
TbExplain (GradCAM) & - & - & 77.33\% \\ \hline
\end{tabular}%
}
\caption{Places365 dataset.}
\label{tab:tab2}
\end{subtable}

\bigskip 

\begin{subtable}{.48\textwidth}
\centering
\resizebox{\linewidth}{!}{%
\begin{tabular}{c|ccc}
\hline
\textbf{Method} & \textbf{Training} & \textbf{Validation} & \textbf{Testing} \\ \hline
ResNet50 & 100\% & 73.13\% & 63.11\% \\
ResNet50V2 & 100\% & 74.62\% & 68.88\% \\
ResNet101 & 100\% & 67.16\% & 65.33\% \\
ResNet101V2 & 100\% & 79.10\% & 79.11\% \\
\textbf{TbExplain (LIME)} & \textbf{-} & \textbf{-} & \textbf{80\%} \\
TbExplain (GradCAM) & - & - & 79.55\% \\ \hline
\end{tabular}%
}
\caption{SUN397 dataset.}
\label{tab:tab3}
\end{subtable}
\end{table}

\section{Conclusion}
\label{conclusion}
In this paper, we propose a novel framework, called TbExplain, that generates text-based explanations for scene classification models. We further attempted to improve classification accuracy by devising a statistical-based approach, called Statistical Prediction Correction (SPC), to correct the prediction by incorporating the confidence of the prediction and scene objects. In the qualitative analysis, we evaluate the reliability and trustworthiness of TbExplain's text-based explanations by comparing them with the manually annotated data labels. Furthermore, the quantitative analysis conducted on three datasets confirmed the efficacy of our proposed framework in improving classification performance.


\bibliographystyle{ACM-Reference-Format}
\bibliography{references}


\end{document}